\documentclass[11pt]{article}
\usepackage[final]{acl}
\usepackage{times}
\usepackage{latexsym}
\usepackage[T1]{fontenc}
\usepackage[utf8]{inputenc}
\usepackage{microtype}
\usepackage{inconsolata}
\usepackage{graphicx}

\usepackage{amsthm}
\usepackage{amsmath}
\usepackage{amsfonts}
\usepackage{amssymb}
\usepackage{autobreak}
\allowdisplaybreaks
\usepackage{adjustbox}
\usepackage{booktabs}
\usepackage{makecell}
\usepackage{siunitx}
\usepackage{multirow}
\usepackage{balance}
\usepackage{longtable}
\usepackage{algorithm}
\usepackage[algo2e]{algorithm2e}
\usepackage{algorithmic}
\usepackage{bm}
\usepackage{enumitem}

\usepackage{graphicx}
\usepackage{placeins}

\usepackage{hyperref}
\usepackage{xcolor}
\usepackage{caption}

\newcommand{\para}[1]{\vspace{2pt}\noindent\textbf{{#1}}}
\def\ourmethod{CatShift}

\title{Hey, That's My Data! Token-Only Dataset Inference in Large Language Models}

\author{
  \textbf{Chen Xiong\textsuperscript{1}},
  \textbf{Zihao Wang\textsuperscript{2}},
  \textbf{Rui Zhu\textsuperscript{3}},\\
  \textbf{Tsung-Yi Ho\textsuperscript{1}},
  \textbf{Pin-Yu Chen\textsuperscript{4}},
  \textbf{Jingwei Xiong\textsuperscript{5}},
  \textbf{Haixu Tang\textsuperscript{2}},
\\
\\
  \textsuperscript{1}The Chinese University of Hong Kong \quad
  \textsuperscript{2}Indiana University Bloomington \quad\\
  \textsuperscript{3}Yale University, School of Medicine, \quad
  \textsuperscript{4}IBM Research\quad
  \textsuperscript{5}University of California, Davis
\\
}

\begin{document}
\maketitle
\begin{abstract}
Large Language Models (LLMs) rely on massive training datasets, often including proprietary data, which raises concerns about unauthorized usage and copyright infringement. Existing dataset inference methods typically require access to log probabilities or other internal signals, but many modern LLMs restrict such access, motivating token-only inference approaches.
We propose CatShift, a token-only dataset inference framework based on catastrophic forgetting, where models overwrite prior knowledge when trained on new data. Fine-tuning an LLM on a subset of its training data induces larger output shifts than fine-tuning on unseen data. CatShift compares these shifts against those from a known non-member validation set to infer whether a dataset was included in training. Experiments on both open-source and API-based LLMs show that CatShift remains effective without logit access, enabling practical protection of proprietary datasets.
\end{abstract}

\section{Introduction}

Large Language Models (LLMs) are increasingly trained on massive web-scale corpora, enabling strong performance in language understanding, reasoning, and generation. However, reliance on large and opaque datasets raises a fundamental question: was a specific dataset, potentially copyrighted or proprietary, used during training? Answering this question is crucial for copyright protection, licensing compliance, and accountability in modern LLM deployment.
Concerns over unauthorized data use have intensified with the commercialization of LLMs. Copyrighted material may be included in training data without consent, infringing authors’ rights and causing financial harm~\cite{DBLP:conf/acl/WeiWJ24, DBLP:journals/corr/abs-2402-02333, DBLP:journals/corr/abs-2402-08787}. A prominent example is the lawsuit filed by the New York Times against OpenAI and Microsoft over alleged misuse of its content~\cite{nyt2023lawsuit}. Media organizations invest heavily in curating high-quality datasets and rarely release them publicly, while even open corpora~\cite{DBLP:journals/jmlr/RaffelSRLNMZLL20, DBLP:journals/corr/abs-2101-00027} often restrict usage to research or educational purposes. Detecting whether such datasets were used for training is essential.

Dataset inference~\cite{maini2024llm, DBLP:conf/iclr/MainiYP21, dziedzic2022dataset} aims to determine whether a model was trained on a given dataset, typically by analyzing logits or log-probabilities. While effective in white-box or logit-accessible settings, these methods become impractical when models expose only outputs from \emph{greedy decoding} (i.e., top-1 predictions), a restriction already enforced by commercial systems such as Claude~\cite{anthropic_claude_3_2024}. This limitation motivates \emph{token-only} inference, which relies solely on predicted tokens. Although token-only access is widely available and robust to logit-perturbation defenses~\cite{DBLP:journals/corr/abs-2402-04013}, naïve strategies such as equating membership with good completion accuracy are unreliable, as shown in \autoref{subsec:Effectiveness}. As a result, more principled approaches are needed.

To address this challenge, we observe that fine-tuning can be repurposed not only for adaptation, but also to probe what models implicitly remember. Our approach leverages \emph{catastrophic forgetting} (CF)~\cite{DBLP:journals/corr/abs-2308-08747}, where learning new data overwrites prior knowledge. If a dataset was part of pre-training, fine-tuning on it again amplifies previously encoded patterns, leading to a pronounced shift in outputs (measured as a larger divergence between pre- and post-fine-tuning completions). In contrast, fine-tuning on unseen data induces smaller and less consistent output changes across samples, reflecting the need to learn new representations.

Building on this insight, \ourmethod{} operationalizes the idea using widely available fine-tuning APIs (e.g., OpenAI, Hugging Face, Google Vertex AI, Claude). The candidate dataset is split into train and test subsets, where the model is fine-tuned on the training portion and evaluated on the test portion. We measure output shifts by comparing pre- and post-fine-tuning completions, using similarity metrics (e.g., BERT-based scorers), and calibrate these shifts against a held-out validation set guaranteed to be unseen. Statistical tests such as KS or Mann–Whitney U are then used to determine whether the observed shift is significantly larger for the target dataset, with a low p-value indicating likely membership.

Our method achieves strong and consistent performance across both open-source and API-based language models, with substantial improvements over prior dataset inference baselines. On the Pythia family of models~\cite{DBLP:conf/icml/BidermanSABOHKP23}, \ourmethod{} achieves an AUC of 0.979 on Pythia-410M, outperforming PDD~\cite{DBLP:conf/emnlp/Zhang0GRFC24} by more than 0.49 AUC points and LLM DI~\cite{maini2024llm} by 0.19 AUC points. At a stringent operating point of 5\% false positive rate, \ourmethod{} further maintains high true positive rates, whereas both baselines degrade sharply under token-only access. On GPT-3.5 with the BookMIA dataset~\cite{shi2024detecting}, member books yield $p=6.44\times10^{-5}$ while non-members yield $p=0.711$, demonstrating robust discrimination under token-only access. These results show that dataset inference remains feasible even under highly restricted interfaces, enabling practical copyright protection without access to logits or privileged internal signals.

\para{Contributions.}
\noindent$\bullet$
We present the first systematic study of dataset inference in token-only settings, targeting realistic scenarios where log probabilities are unavailable.

\noindent$\bullet$
We introduce \ourmethod{}, a CF-based inference approach that consistently outperforms existing baselines across models and datasets.

\noindent$\bullet$
We provide empirical evidence that training members exhibit significantly larger output shifts after fine-tuning than non-members, explaining the effectiveness of CF-based inference.

\section{Background}
\label{sec:back}

\subsection{Dataset Inference Attack}
\label{subsec:dia}
\textit{Dataset inference attacks} aim to distinguish data drawn from the model's training distribution $p_{\mathcal{D}_{\text{train}}}$ from data sampled from a non-member distribution $p_{\mathcal{D}_{\text{non-train}}}$. Following Maini et al.~\cite{DBLP:conf/iclr/MainiYP21}, who observed that training data often lies farther from decision boundaries, these attacks exploit distributional shifts in features, logits, or model outputs to infer membership collectively.

Formally, a scoring function $s(x)$ estimates the distance or likelihood of a sample $x$ under $p_{\mathcal{D}_{\text{train}}}$. A threshold $\tau$ is applied: $s(x) > \tau \Rightarrow x \in \mathcal{D}_{\text{train}}$; otherwise, $x \in \mathcal{D}_{\text{non-train}}$. Recent work~\cite{maini2024llm} extends this to LLMs, where fine-tuning induces measurable distribution shifts. Hypothesis testing can then be used for membership inference, but requires labeled member/non-member data and a separate validation set for calibrating $\tau$.

\subsection{Fine-Tuning APIs}
\label{subsec:finetune-api}
Fine-tuning APIs are now widely offered for LLMs, enabling users to adapt powerful pre-trained models to specific domains without full retraining. Platforms like OpenAI's GPT-3.5 and Hugging Face allow developers to fine-tune models on custom data for applications such as chatbots, legal document analysis, and clinical text classification.

Fine-tuning integrates domain-specific knowledge into general-purpose models, improving performance on specialized terminology and tasks. For instance, healthcare models better extract clinical entities, and legal models handle contracts more accurately. Fine-tuning also supports domain adaptation across languages and regions, boosting robustness where generic models underperform.
These APIs provide model customization without exposing internals, lowering technical barriers and enabling adoption in privacy-sensitive industries. Standardized interfaces make advanced LLM capabilities accessible to a wide range of sectors.

\subsection{Threat Model}
\label{subsec:threat-model}

\para{Adversary and Goals.}
We consider an adversarial LLM provider who may have trained on proprietary data without consent. The \emph{dataset owner} seeks to determine whether their data was used in training under restrictive query conditions, even if the provider withholds log probabilities or manipulates outputs to conceal unauthorized usage.

\para{Defender Capabilities.}
The dataset owner has black-box access to the target LLM through a token-only interface, observing only outputs from \emph{greedy decoding} (i.e., top-1 completions). No logits, confidence scores, or token probabilities are exposed. The defender can fine-tune the LLM via publicly available APIs and possesses a validation set that is guaranteed not to be included in the model’s training corpus.
Such non-member data is realistic in our setting, as dataset owners typically have access to private or newly generated content (e.g., internal documents or post-collection data) that is unlikely to appear in pretraining corpora. This non-member set serves as a baseline for measuring output distribution shifts.

\para{Adversarial Constraints.}
The provider may hide membership signals by withholding log probabilities, randomizing outputs, or limiting fine-tuning options (e.g., training duration or capacity). These measures aim to suppress overfitting indicators and reduce the owner's ability to probe the model.

\para{Assumptions and Practical Considerations.}
We assume that fine-tuning genuinely updates model parameters and that pre- and post-fine-tuning queries are evaluated under a fixed model version. Large backend updates (e.g., model refreshes or safety filter changes) could introduce additional variation in outputs. In practice, this can be mitigated by conducting evaluations within a short time window or the same API session, and by calibrating results using the non-member validation set.

\para{Attack Surface and Desired Protections.}
Under these conditions, the dataset owner's strategy is to detect unauthorized data use through token-only queries and statistical comparison of fine-tuning effects on candidate versus validation datasets. This enables robust dataset inference even under restricted interfaces and limited signal exposure.

\section{Methodology of \ourmethod{}}
\label{sec:method}

\subsection{Design Motivation}
We briefly recap the key intuition introduced in the Introduction. Under token-only access, conventional dataset inference signals (e.g., likelihoods) are unavailable, making membership detection challenging.

Our approach leverages \emph{catastrophic forgetting}: fine-tuning on previously seen data reactivates latent representations and induces larger output shifts, while unseen data leads to smaller, less structured changes. This asymmetry enables distinguishing member from non-member datasets through output-level comparisons.

\begin{figure*}[!t]
\centerline{\includegraphics[width=\linewidth]{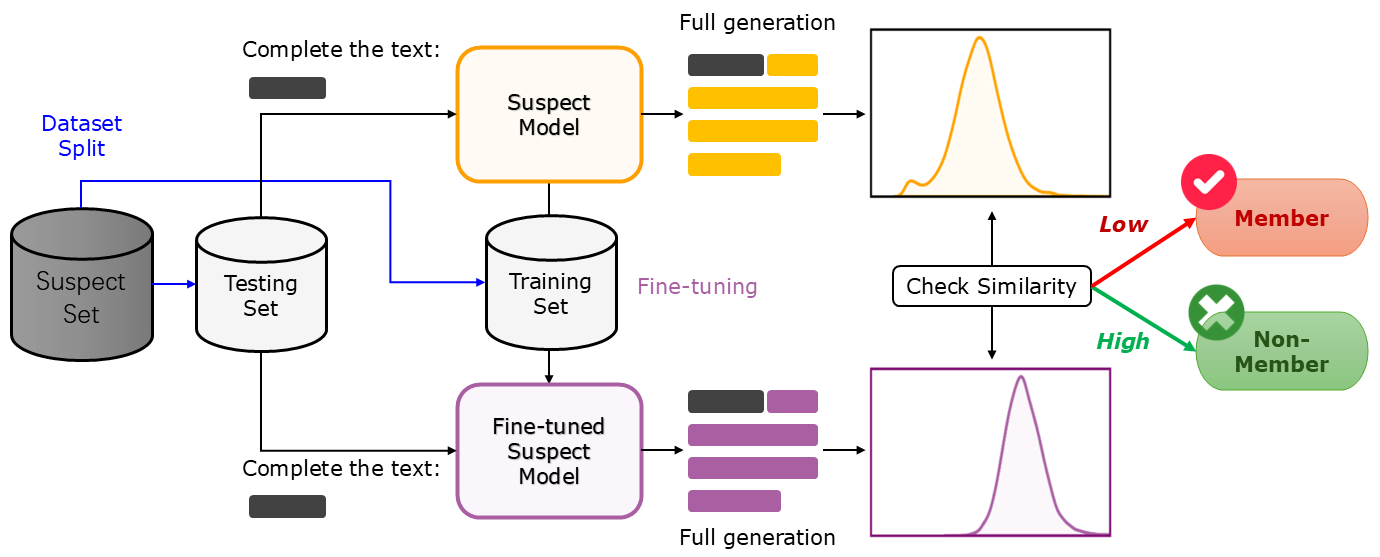}}
\caption{\small An overview of the \ourmethod{} framework for dataset inference on token-only large language models (LLMs). The figure illustrates the three-step process: (1) Dataset Split; (2) Completion Prompt Construction; (3) Target Model Fine-Tuning; (4) Output Distribution Analysis.}
\label{fig:overview}
\end{figure*}

\subsection{Overview of \ourmethod{}}
\label{sec:method_overview}

To address the token-only constraint, \ourmethod{} combines \emph{catastrophic forgetting} with \emph{statistical hypothesis testing} to distinguish a target dataset from a known non-member validation set, without relying on confidence scores or internal states. The key insight is that fine-tuning affects member and non-member data asymmetrically at the output level, even as model parameters drift away from their pre-trained configuration.

Fine-tuning induces parameter drift characteristic of catastrophic forgetting. For member data, this process re-optimizes around latent representations already encoded during pre-training, leading to comparatively rigid or structured shifts in token-level outputs. In contrast, non-member data requires forming new representations, resulting in more diffuse and less stable output changes. This asymmetry yields a detectable statistical signal under token-only observation, as shown in Fig.~\ref{fig:overview}.

Guided by this observation, \ourmethod{} measures output distribution shifts induced by controlled fine-tuning and compares them against those observed on a held-out non-member dataset.

\para{Step 1: Completion Prompt Construction.}
We frame dataset inference as a text-completion task. Each sample is split into a prompt and completion, and the prompt is placed inside a generic instruction (e.g., ``Complete the following text''). This setup works with token-only interfaces since it requires only top-1 predictions, aligning with typical commercial LLM APIs.

\para{Step 2: Target Model Fine-Tuning.}
We fine-tune the target LLM on these prompt–completion pairs. If the dataset was part of the training corpus, fine-tuning rapidly refreshes internal representations, causing a pronounced shift in outputs. For unseen data, the model learns new patterns more gradually, producing smaller shifts.

\para{Step 3: Output Distribution Analysis.}
We collect pre- and post-fine-tuning completions, compute similarity metrics (e.g., lexical overlap or BERT-based scores), and compare them with those from the validation set. Statistical tests determine whether the target dataset's shift is significantly larger, signaling likely membership. Because this process relies only on top-1 predictions, it remains robust even when logits or probabilities are hidden.

\subsection{Completion Prompt Construction}  
\label{subsec:prompt_construction}

\noindent
We represent each sample in the \emph{target dataset} as a prompt–completion pair:
\begin{equation}
D_s = {(x_i, y_i)}_{i=1}^{N},
\end{equation}
where $x_i$ is the prompt and $y_i$ the completion. In practice, we split each document in half, using the first half as the prompt and the second as the completion, ensuring consistency across samples. Under the token-only setting, the LLM produces a continuation $\hat{y}_i$ for each $x_i$, which we record for later analysis. This construction is fully compatible with standard LLM APIs and enables dataset owners to test for unauthorized data usage.

If the model is instruction-tuned, we directly use prompts to elicit completions, leveraging its ability to follow explicit instructions for more reliable outputs. Otherwise, we use simple prefixes to trigger causal continuations. Since models tend to assign higher likelihood to training data, prefixes from member data often lead the model to reproduce it, making prefix-based continuation a practical strategy for eliciting memorized content in non-instruction-tuned models.

\subsection{Target Model Fine-Tuning}
\label{subsec:fine_tuning}

\noindent
We fine-tune the target LLM $f^{(0)}$ with parameters $\theta^{(0)}$ on a subset of the target dataset. The dataset is split into $D_s^{\text{train}}$ and $D_s^{\text{test}}$ with no overlap, and fine-tuning updates the parameters to $\theta^{(1)}$ by minimizing

$$
\theta^{(1)} = \underset{\theta}{\mathrm{arg\,min}} \; \sum_{(x_i, y_i) \in D_s^{\text{train}}} \mathcal{L}\bigl(f^{(0)}(x_i;\theta), y_i\bigr),
$$

where $\mathcal{L}$ is typically cross-entropy loss. If $D_s$ was part of the original training data, re-exposure refreshes its patterns and produces a stronger change in predictions; if it is novel, the model must learn new patterns, resulting in more gradual shifts. This distinction underpins our later statistical test.

To perform fine-tuning efficiently, we use Low-Rank Adaptation (LoRA)~\cite{DBLP:conf/iclr/HuSWALWWC22}, which updates a low-rank decomposition of weight matrices, reducing computational cost while preserving performance. We apply LoRA to the Query-Key-Value (QKV) attention layer, the linear layer following attention, and the MLP layer. Fine-tuning these components allows the model to adjust its attention weights, refine internal representations, and update prediction mappings, all while retaining most pretrained parameters.

By focusing adaptation on these layers, we achieve rapid, parameter-efficient fine-tuning that highlights differences between reactivating existing knowledge and learning from scratch. Member datasets trigger larger shifts in top-1 predictions, whereas non-member datasets produce subtler changes—providing a measurable signal for dataset membership inference.

\subsection{Output Distribution Analysis}
\label{subsec:output_analysis}

\noindent
We quantify the \emph{shift} in outputs by comparing completions before and after fine-tuning. For each $(x_i, y_i) \in D_s^{\text{test}}$, we collect

$$
\hat{y}_i^{(0)} = f^{(0)}(x_i), \qquad
\hat{y}_i^{(1)} = f^{(1)}(x_i),
$$

where $\hat{y}_i^{(0)}$ and $\hat{y}_i^{(1)}$ are top-1 completions from the original and fine-tuned models.

\noindent
\para{Similarity Scores.}\quad
We measure the change with a similarity function $\mathrm{Sim}(\cdot,\cdot)$ (e.g., n-gram overlap or BERT score):

$$
s_i = \mathrm{Sim}\!\bigl(\hat{y}_i^{(0)}, \hat{y}_i^{(1)}\bigr),
$$

where smaller $s_i$ indicates greater divergence and thus a larger shift.

\noindent
\para{Validation Baseline.}\quad
To calibrate what constitutes a “normal” shift, we repeat the same process on a known non-member validation set $D_v = \{(x_j^v, y_j^v)\}_{j=1}^{M}$, obtaining

$$
s_j^v = \mathrm{Sim}\!\bigl(f^{(0)}(x_j^v),\, f^{(1)}(x_j^v)\bigr)
$$

This yields a baseline distribution of shifts for non-member data.

\noindent
\para{Hypothesis Testing.}\quad
Let $S_s = \{s_i\}$ and $S_v = \{s_j^v\}$ be the similarity score sets for the target and validation data. We apply a two-sample test such as Kolmogorov–Smirnov:

$$
p\text{-value} = \mathrm{KS}(S_s, S_v).
$$

The p-value measures the statistical evidence that the target dataset induces a larger shift than the non-member baseline. It is used as a decision statistic rather than a performance score: if $p\text{-value} < \alpha$ (e.g., $\alpha = 0.1$), we reject the null hypothesis and infer likely membership. Different datasets may yield different absolute p-values due to heterogeneity, but the decision depends on whether the threshold is crossed while controlling the false positive rate.

By relying solely on token-only completions and a robust validation baseline, \ourmethod{} detects unauthorized data usage even when log probabilities are withheld or obfuscated, making it practical for real-world LLMs.

\para{Why does CatShift work?}
We provide a brief theoretical intuition and defer detailed discussion to \autoref{appendix:theory_catshift}. Our reasoning builds on two observations from prior work. First, memorized or training samples are often associated with sharper local geometry in the loss landscape, implying that parameter updates along these directions lead to disproportionately larger functional changes. Second, prior work on membership inference has consistently identified behavioral asymmetry between training and non-training data (e.g., in confidence and loss), suggesting that models respond differently around member versus non-member regions.

Under these conditions, fine-tuning on member data perturbs the model along more sensitive directions, resulting in larger and more consistent output shifts compared to non-member data. In contrast, non-member data typically induces weaker and less structured changes. We also rule out alternative explanations such as token-frequency effects in \autoref{appendix:theory_catshift}.

\para{Overall Algorithm.}
The full pseudocode of the proposed \ourmethod{} framework is provided in 
\autoref{appendix:overall_algorithm} for completeness. 
Here, we focus on the intuition and main components that drive the framework's design.

\section{Evaluation}
\label{sec:eval}

\subsection{Experimental Setup}
\label{subsec:setup}

\para{Datasets and Models.}
We evaluate on the full PILE dataset~\cite{DBLP:journals/corr/abs-2101-00027}, which comprises 22 subsets, using both public checkpoints and API-based models.
For public models, we use Pythia~\cite{DBLP:conf/icml/BidermanSABOHKP23} and GPT-Neo~\cite{black2021gptneo}, both trained on PILE. Because the training and validation splits of all 22 subsets are public, we can precisely label members (training samples) and non-members (validation samples) under IID conditions with no temporal shift, using each subset's validation split as its baseline.
For API-based evaluation, we fine-tune GPT-3.5 through OpenAI's API and use the BookMIA dataset~\cite{shi2024detecting}, which contains known member and non-member books for OpenAI models.
Unless stated otherwise, we sample 1000 training and 1000 validation examples per PILE subset for membership evaluation, and use an additional 600 samples for fine-tuning in \ourmethod{}.

We adhered to established ethical guidelines. Most experiments used publicly available datasets in a controlled local environment. For GPT-3.5, we fine-tuned using only public data and did not release the resulting model, mitigating misuse risks. As no human participants were involved, our IRB determined this work does not constitute human-subjects research. While misuse is possible, for example if malicious actors use dataset-inference techniques to probe private models, we disclose these findings to raise awareness and encourage constructive discussion.

Note that we used LLMs only to polish the writing style (e.g., grammar and spelling).

\para{Baselines.}
We compare \ourmethod{} against two representative dataset inference baselines for large language models: LLM DI~\cite{maini2024llm} and PDD~\cite{DBLP:conf/emnlp/Zhang0GRFC24}.

LLM DI~\cite{maini2024llm} formulates dataset inference as a statistical testing problem based on differences in model loss or likelihood statistics between candidate and reference datasets. While effective when logits or loss values are accessible, it fundamentally relies on probabilistic outputs, which are unavailable under token-only interfaces.

PDD~\cite{DBLP:conf/emnlp/Zhang0GRFC24} detects pretraining data by measuring calibrated divergences between output distributions induced by controlled perturbations. This approach similarly assumes access to probability distributions or calibrated likelihood estimates to compute divergence scores, limiting its applicability in strict black-box settings where only top-1 completions are exposed.

In contrast, \ourmethod{} operates solely on top-1 completions and does not require access to logits, probabilities, or loss values, making it compatible with LLM APIs that restrict output visibility.

\para{Evaluation Protocol.}
We evaluate \ourmethod{} on each of the 22 PILE subsets, treating training splits as members and validation splits as non-members. For each subset, we apply the Kolmogorov–Smirnov (KS) test to compare the distributions of similarity scores obtained from candidate data and a non-member validation set:
\[
D = \sup_x \lvert F_n(x) - F_m(x) \rvert .
\]
The resulting p-values serve as continuous membership scores, with smaller values indicating stronger evidence of membership. Rather than relying on a single decision threshold, we evaluate performance using threshold-independent metrics, reporting ROC AUC to measure ranking quality and TPR at a fixed 5\% false positive rate (TPR@5\%) to assess sensitivity under conservative operating conditions relevant for practical auditing.

\para{\ourmethod{} Configuration.}
We fine-tune with 600 samples per subset. For GPT-3.5, we adopt OpenAI's default configuration; for Pythia and GPT-Neo, we use LoRA~\cite{DBLP:conf/iclr/HuSWALWWC22} with rank 8, $\alpha=32$, and dropout 0.1, training with learning rate $8 \times 10^{-5}$ and batch size 8. Checkpoints are saved every 10 steps, and the model with the lowest normalized combined loss (on both member and non-member sets) is selected. Fine-tuning and test sets are strictly disjoint.

\para{Computational Cost.}
Although our method relies on fine-tuning, its computational cost remains modest in practice. Dataset inference requires only a small number of fine-tuning iterations on a limited subset of samples, rather than full retraining. In our experiments, performing detection with GPT-3.5 incurs an estimated cost of approximately \$156 per target dataset, well within the budget of routine auditing or compliance checks. This demonstrates that fine-tuning-based dataset inference is not prohibitively expensive, and can be realistically deployed for practical ownership verification and misuse detection.

\begin{table}[!t]
\small
\caption{Dataset inference performance across models under token-only access. We compare CatShift against two representative baselines, PDD and LLM DI, using AUC and TPR at 5\% false positive rate (TPR@5\%).}

\resizebox{0.48\textwidth}{!}{
\begin{tabular}{cccc}
\toprule
Model & Method & AUC & TPR@5\% \\
\midrule
\multirow{3}{*}{\begin{tabular}[c]{@{}c@{}}Pythia\\ 410M\end{tabular}} & PDD~\cite{DBLP:conf/emnlp/Zhang0GRFC24} & 0.4865 & 0.0499 \\
 & LLM DI~\cite{maini2024llm} & 0.7890 & 0.5910 \\
 & CatShift (Ours) & \textbf{0.9790} & \textbf{0.9545} \\
 \midrule
\multirow{3}{*}{\begin{tabular}[c]{@{}c@{}}GPT-Neo\\ 125M\end{tabular}} & PDD~\cite{DBLP:conf/emnlp/Zhang0GRFC24} & 0.5000 & 0.0680 \\
 & LLM DI~\cite{maini2024llm} & 0.9122 & 0.6860 \\
 & CatShift (Ours) & \textbf{0.9940} & \textbf{0.9561} \\
 \midrule
\multirow{3}{*}{\begin{tabular}[c]{@{}c@{}}GPT-Neo\\ 1.3B\end{tabular}} & PDD~\cite{DBLP:conf/emnlp/Zhang0GRFC24} & 0.4896 & 0.0507 \\
 & LLM DI~\cite{maini2024llm} & 0.8200 & 0.4113 \\
 & CatShift (Ours) & \textbf{0.9410} & \textbf{0.8227} \\
 \midrule
\multirow{3}{*}{\begin{tabular}[c]{@{}c@{}}GPT-NeoX\\ 20B\end{tabular}} & PDD~\cite{DBLP:conf/emnlp/Zhang0GRFC24} & 0.5574 & 0.0731 \\
 & LLM DI~\cite{maini2024llm} & 0.9280 & 0.7727 \\
 & CatShift (Ours) & \textbf{1.0000} & \textbf{0.9640} \\
 \bottomrule
\end{tabular}
\label{tab:main}
}
\end{table}

\begin{figure*}[!t]
\centerline{\includegraphics[width=\linewidth]{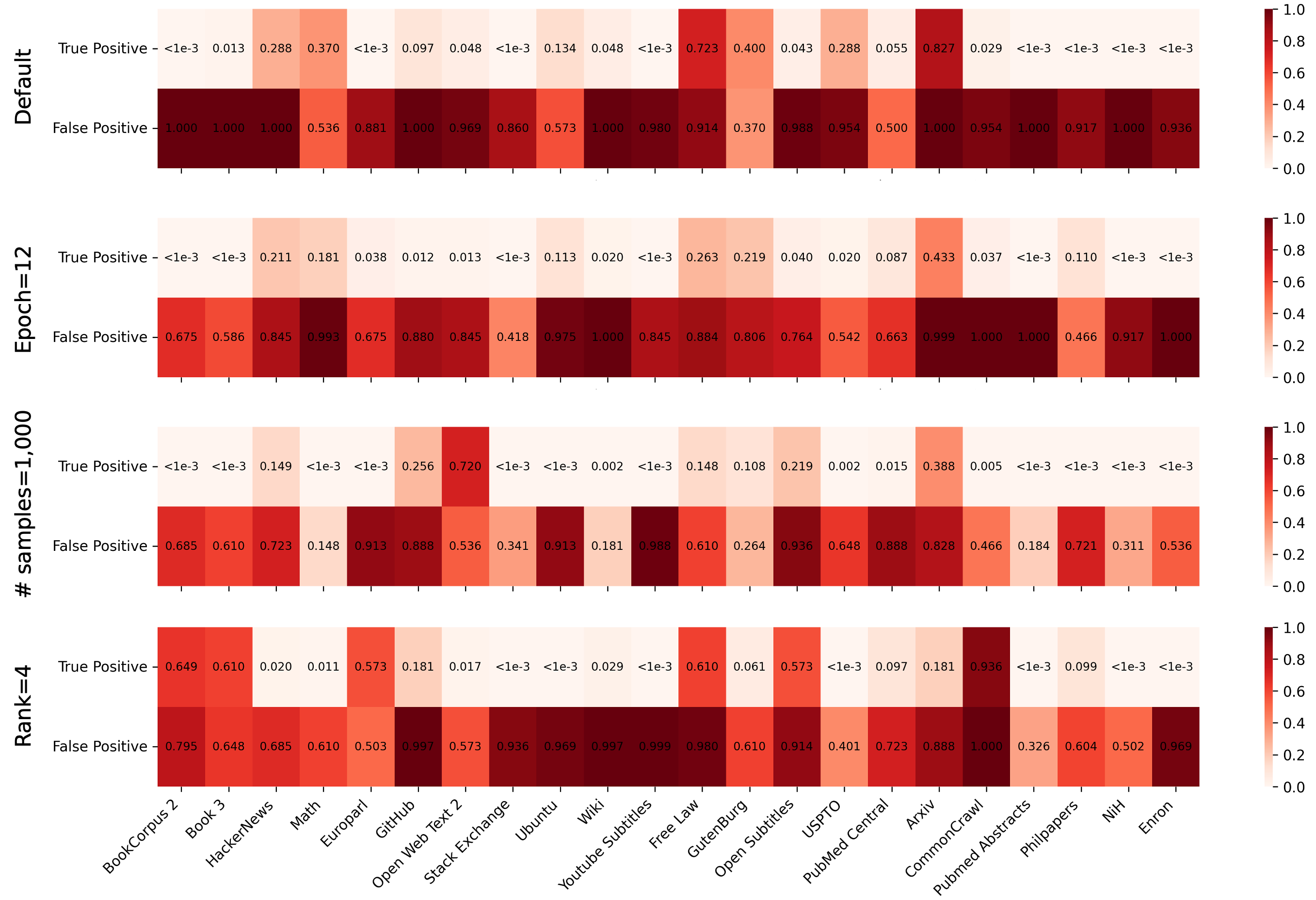}}
\caption{Ablation study. p-values of dataset inference on the Pythia-410M model under different ablation settings. We vary the fine-tuning duration by increasing the number of epochs to 12 and modify the LoRA configuration by reducing the rank to 4, while keeping all other settings unchanged.}
\label{fig:ablation}
\end{figure*}

\subsection{Effectiveness of \ourmethod{}}
\label{subsec:Effectiveness}

We evaluate the effectiveness of \ourmethod{} under token-only access across multiple model architectures and scales, comparing it against two representative dataset inference baselines: PDD and LLM DI. Both baselines are evaluated under their applicable access assumptions, while \ourmethod{} relies exclusively on top-1 completions.

\autoref{tab:main} reports AUC and TPR at a fixed 5\% false positive rate (TPR@5\%), a stringent operating point relevant for practical auditing scenarios. Across all models, CatShift consistently achieves the highest AUC, indicating superior ranking quality between member and non-member datasets independent of any decision threshold. On Pythia-410M, CatShift attains an AUC of 0.979, substantially outperforming LLM DI (0.789) and PDD (0.487). Similar trends hold for GPT-Neo models at both 125M and 1.3B scales, as well as the 20B GPT-NeoX model, where CatShift achieves near-perfect discrimination.

Performance gaps become even more pronounced when considering low false positive regimes. At 5\% FPR, CatShift maintains TPRs above 0.95 on Pythia-410M and GPT-Neo 125M, and above 0.82 even on GPT-Neo 1.3B. In contrast, PDD exhibits near-random behavior across all settings, with TPR@5\% close to chance level, while LLM DI shows moderate sensitivity but degrades notably on larger models. These results indicate that probability- and divergence-based methods fail to reliably detect membership when restricted to token-level signals.

Overall, CatShift demonstrates both high sensitivity and strong robustness under conservative operating conditions. It consistently detects a large fraction of true members while tightly controlling false positives, and its advantage persists across model families and scales. These findings confirm that catastrophic-forgetting-induced output shifts provide a reliable and scalable signal for dataset inference in logit-inaccessible settings.

Detailed analyses across subsets and failure-case analyses are presented in \autoref{subsec:Model}.

\subsection{Effectiveness on Commercial Models}
\label{subsec:API}

In this section, we investigate the effectiveness of \ourmethod{} on API-based commercial models, specifically focusing on OpenAI's GPT-3.5. We utilize the fine-tuning interface provided by OpenAI, which is designed to adapt the model to domain-specific tasks by fine-tuning it on custom datasets. For this evaluation, we use the default fine-tuning settings offered by OpenAI, ensuring that the model is trained according to the standard procedure.

However, since the membership status of these models is not publicly disclosed, we refrain from using the PILE dataset for evaluation, as we did in previous sections. Instead, we turn to the BookMIA dataset~\cite{shi2024detecting}, which is specifically designed for evaluating membership status. The BookMIA dataset contains both member and non-member books from OpenAI models, making it an ideal candidate for assessing the membership prediction capabilities of \ourmethod{}.

The results show that \ourmethod{} achieves a p-value of \(6.44 \times 10^{-5}\) for member books, indicating strong evidence that these books belong to the model's training data. In contrast, the non-member books have a p-value of 0.711, which is significantly higher, demonstrating that \ourmethod{} effectively distinguishes between member and non-member data. This clear separation in p-values highlights the robustness and applicability of \ourmethod{} when applied to commercial models, such as those offered by OpenAI, further validating its effectiveness in real-world scenarios.

\subsection{Ablation Study}

We examine the sensitivity of \ourmethod{} to key hyperparameters that influence fine-tuning dynamics, focusing on the number of fine-tuning epochs and the LoRA rank.

\para{Impact of Fine-Tuning Epochs.}
We evaluate the effect of training duration by comparing the default setting of 9 fine-tuning epochs with an extended setting of 12 epochs on Pythia-410M across all 22 PILE subsets. As shown in \autoref{fig:ablation}, at a threshold of $p < 0.1$, the epoch=9 configuration correctly identifies 15 member datasets, while epoch=12 identifies 14. Importantly, both settings yield zero false positives, preserving perfect specificity on non-member subsets. These results indicate that modest variations in fine-tuning duration have limited impact on detection performance, and that additional epochs do not necessarily improve sensitivity.

\para{Impact of Fine-Tuning Sample Size.}
We further examine the sensitivity to the size of the fine-tuning dataset by varying the number of samples from 600 to 1,000 on Pythia-410M across all 22 PILE subsets. As shown in \autoref{fig:ablation}, the overall detection performance remains largely stable under this change. Increasing the number of samples does not consistently strengthen true-member detection: subsets with strong signals remain highly significant in both settings, while others exhibit only minor fluctuations. At the same time, a larger sample size generally improves stability on non-member subsets, leading to reduced false positives. This trend is also reflected in the overall AUC, which changes only slightly from 0.979 (600 samples) to 0.945 (1000 samples), indicating modest variation rather than structural degradation. Overall, the number of detected true positives remains comparable across both settings, suggesting that \ourmethod{} does not rely on large fine-tuning datasets and is robust to sample-size variation.

\para{Impact of LoRA Rank.}
We next study the effect of parameter capacity by varying the LoRA rank, comparing the default rank=8 with a reduced rank=4 under otherwise identical conditions. As illustrated in \autoref{fig:ablation}, rank=8 correctly detects 15 member datasets, while rank=4 detects 14 at the same $p < 0.1$ threshold. Both configurations again produce no false positives. This suggests that the method does not rely on aggressive parameter updates and remains effective with lightweight fine-tuning.

Overall, the ablation results indicate that \ourmethod{} is not highly sensitive to fine-tuning hyperparameters. Neither increasing the number of epochs nor reducing the LoRA rank leads to meaningful changes in detection accuracy. This suggests that the signal exploited by \ourmethod{} stems from relative output shifts caused by catastrophic forgetting, which emerge even under lightweight and conservative fine-tuning. As a result, the method remains effective without careful hyperparameter tuning, making it suitable for practical deployment.

\begin{figure}[!ht]
\centerline{\includegraphics[width=.9\linewidth]{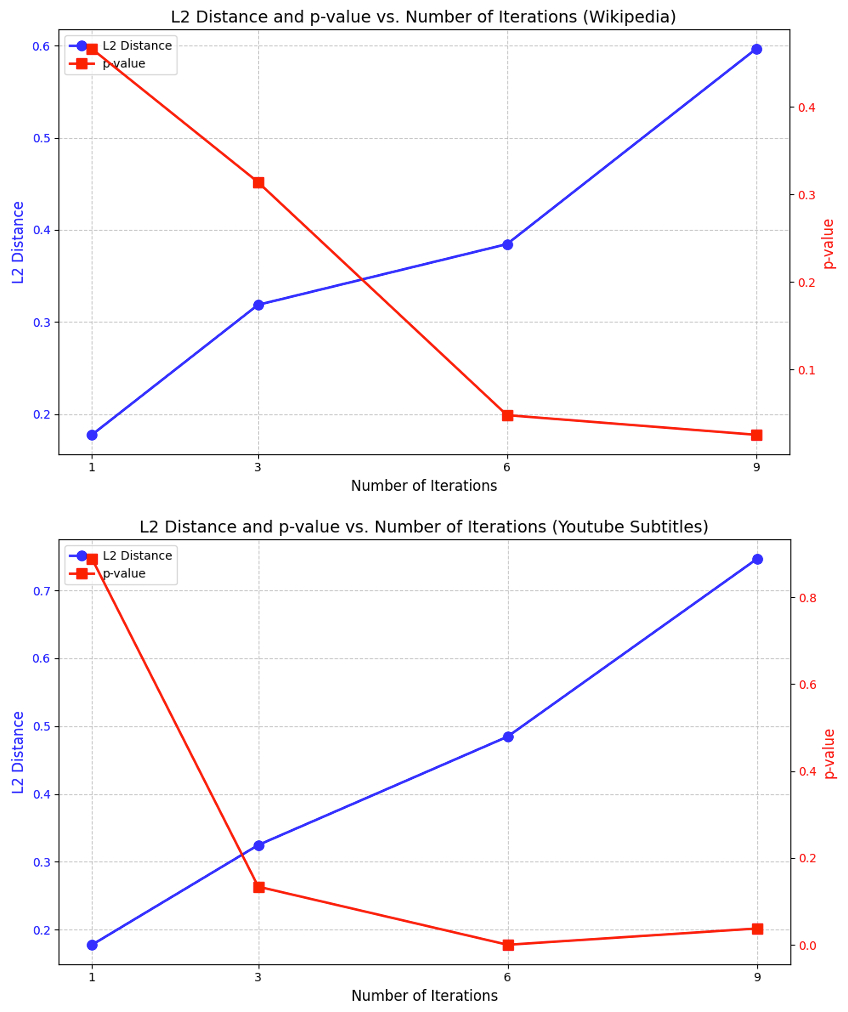}}
\caption{Effect of fine-tuning iterations on \ourmethod{}.}
\label{fig:iteration}
\end{figure}

\subsection{Impact of Fine-Tuning Iterations}
\label{subsec:iteration}
In this section, we study how the number of fine-tuning iterations affects \ourmethod{}. \autoref{fig:iteration} shows results on the Wikipedia and YouTube Subtitles subsets.
As iterations increase, the L2 distance between fine-tuned and original weights grows monotonically, capturing the progressive parameter drift characteristic of catastrophic forgetting. Continued optimization overwrites pre-trained representations, fitting the fine-tuning data more closely. At the same time, p-values from our hypothesis test decrease, indicating that the model's output distribution becomes increasingly distinguishable from that of a non-member dataset. This supports our hypothesis that fine-tuning on member data not only learns new patterns but also reactivates latent knowledge, producing significant shifts in top-1 completions.
The inverse relationship between L2 distance and p-value highlights how parameter-level forgetting amplifies output-level membership signals. Our results confirm that the more a model ``forgets'' during adaptation to member data, the more clearly it ``remembers'' it, yielding a strong statistical signature for dataset inference.

\subsection{Robustness to Perturbations}
\label{subsec:robustness}

CatShift relies on the standard API assumption that fine-tuning updates model parameters and that outputs reflect the current model state. Under this setting, we examine the robustness of CatShift to potential obfuscation strategies.

For parameter perturbations, prior work on catastrophic forgetting and selective unlearning~\cite{seam} suggests that moderate weight noise applied before fine-tuning is unlikely to eliminate structured signals, as subsequent fine-tuning re-optimizes the model toward the target dataset. In contrast, suppressing detection via post-fine-tuning perturbations would require sufficiently large changes to alter top-1 outputs across many inputs, which also affects model behavior.

We further evaluate output perturbation by injecting Gaussian noise into the logits of a fine-tuned Pythia-410M model (YouTube Subtitles subset of The Pile). Results are shown in \autoref{tab:perturb}.

\begin{table}[h]
\centering
\small
\begin{tabular}{ccc}
\toprule
\textbf{Perturbation Level} & \textbf{P-Value} & \textbf{Perplexity} \\
\midrule
0.0 & $<10^{-3}$ & 22.915 \\
0.5 & $<10^{-3}$ & 24.109 \\
1.0 & 0.042 & 24.603 \\
1.5 & 0.088 & 24.902 \\
2.0 & 0.248 & 54.517 \\
\bottomrule
\end{tabular}
\caption{Effect of output perturbation on detection and model quality.}
\label{tab:perturb}
\end{table}

Moderate perturbations (0.5–1.5) have limited impact on detection, with p-values remaining below the threshold (0.1), while large perturbations (2.0) suppress detection but significantly degrade perplexity. These results indicate a trade-off between obfuscating membership signals and preserving model utility.

\section{Related Work}
\label{sec:related} 
MIA techniques, first developed for classification models~\cite{MIA}, have recently been adapted for LLMs. Likelihood-based metrics such as loss~\cite{Overfitting} and perplexity~\cite{Extracting_LLM} are commonly used to distinguish training members from non-members, often yielding high AUC-ROC scores. Min-k\%Prob~\cite{shi2024detecting} extends this idea by averaging the log-likelihood of the lowest-probability $k\%$ tokens in a document.
Neighbor-based MIAs refine perplexity scores by comparing the likelihoods across neighboring models $M'$ or documents $D'$~\cite{Duan2024-doMIAwork, galli2024noisy}. While $M'$-based approaches require a reference model trained on a disjoint but similar dataset—often unrealistic—$D'$-based methods are more practical but generally less effective and sensitive to noise parameters.

Recent work~\cite{Duan2024-doMIAwork, maini2024llm} highlights biases that inflate reported MIA performance, such as temporal shifts between member and non-member data. Re-evaluation on Pythia~\cite{DBLP:conf/icml/BidermanSABOHKP23}, which uses unbiased random splits of The Pile~\cite{DBLP:journals/corr/abs-2101-00027}, shows significantly lower AUC-ROC scores. Moreover, simple classifiers can exploit these biases to separate members from non-members~\cite{das2024blind, meeus2024inherent}, underscoring the need to evaluate MIAs on truly random datasets.

Such evaluation is feasible for open models like Pythia but not for proprietary LLMs that do not disclose training data (e.g., those in copyright disputes). As observed in multiple studies~\cite{Duan2024-doMIAwork, maini2024llm, meeus2024copyright}, MIA performance varies widely across models and datasets, so results from open LLMs cannot be directly generalized to closed ones.

\section{Conclusion}
\label{sec:conclusion}
In this paper, we introduced \ourmethod{}, a token-only dataset-inference framework that exploits catastrophic forgetting to determine whether a dataset was used in training a large language model. The method fine-tunes the model on part of the suspicious dataset and compares output shifts against those induced by a non-member validation set, enabling reliable detection of proprietary data usage even when log probabilities are inaccessible.
Extensive experiments on both open-source and commercial LLMs show that \ourmethod{} consistently reduces false positives and false negatives under token-only conditions. These results underscore its practicality for auditing and compliance in real-world scenarios. At the same time, the observed variation across model families highlights opportunities for further study on how architecture and scale influence catastrophic forgetting, pointing to promising directions for advancing token-only dataset inference.

\section*{Limitations}

Despite the strengths of \ourmethod{}, several practical constraints and open questions remain. First, catastrophic forgetting may be weaker or more difficult to detect for certain classes of datasets, especially those that overlap substantially with other data in the model's original training corpus. In such scenarios, it may be challenging to disentangle shifts caused by \emph{true reactivation} of learned knowledge from those arising due to partial redundancy with content already present in the model's parameters. Second, the fine-tuning procedure itself introduces hyperparameters (e.g., learning rate, number of epochs, or choice of layers to update) that can significantly affect the observed output shifts. Determining optimal hyperparameter configurations for membership inference, especially under limited budget constraints, remains an open research question. Finally, while our token-only approach targets modern APIs where log probabilities are withheld, providers may deploy additional strategies (e.g., heavy output truncation or specialized filters) that can further limit the information available to dataset owners, thereby complicating or weakening the method's efficacy.

\section*{Acknowledgment and Funding Statement}

Chen Xiong and Tsung-Yi Ho, from the JC STEM Lab of Intelligent Design Automation, are funded by the Hong Kong Jockey Club Charities Trust.


\appendix

\section{Theoretical Intuition and Conditions for CatShift}
\label{appendix:theory_catshift}

We provide additional theoretical intuition to explain when and why \ourmethod{} is effective. Our analysis builds on two well-supported observations from prior work, followed by a discussion of potential confounding factors and the conditions under which the method succeeds or fails.

\subsection{Curvature and Sensitivity of Memorized Data}

A growing body of work suggests that memorized or training examples are associated with sharper local geometry and higher curvature in the loss landscape. Recent studies~\cite{garg2023memorization, ravikumar2024unveiling} show that per-example loss curvature correlates strongly with memorization, and that memorized samples tend to lie in sharper regions compared to non-memorized data. These results indicate that curvature-related quantities can serve as reliable signals of memorization.

In the context of CatShift, this implies that fine-tuning on member data perturbs model parameters along more sensitive and high-curvature directions, leading to disproportionately larger functional changes. In contrast, non-member data typically lies in flatter regions, where parameter updates induce weaker effects on model outputs.

\subsection{Behavioral Asymmetry Between Members and Non-Members}

Another key assumption in membership and dataset inference literature is that models exhibit systematic behavioral asymmetry between training and non-training data. Prior work~\cite{DBLP:conf/iclr/MainiYP21, chen2023overconfidence} shows that training samples tend to receive higher confidence scores, lower prediction loss, and larger margins to decision boundaries compared to non-members.

This asymmetry implies that the model's response surface differs around member versus non-member regions. CatShift leverages this property at the functional level by measuring differences in model outputs before and after controlled fine-tuning, effectively capturing this asymmetry without requiring access to logits or confidence scores.

\subsection{Ruling Out Token-Frequency Confounds}

A potential concern is that the observed output shifts are driven by superficial token-frequency effects rather than membership-specific signals. We mitigate this confound in two ways.

First, shifts are measured on a held-out test split that is strictly disjoint from the fine-tuning split, ensuring that the signal cannot be explained by memorizing fine-tuning samples.

Second, in our PILE setting, members (training split) and non-members (validation split) are independently and identically distributed (IID) samples from the same subset. As a result, their token-frequency and stylistic statistics are closely matched, reducing the likelihood that frequency-based artifacts drive the observed differences.

Moreover, \autoref{subsec:iteration} shows that as fine-tuning iterations increase, the $\ell_2$ distance between the fine-tuned and original model parameters grows monotonically, while the KS-test p-value decreases. This indicates that the relative shift between candidate and non-member baselines becomes increasingly significant as parameter drift increases.

If the effect were primarily driven by generic overfitting to frequent tokens, we would expect comparable drift-induced shifts on both IID splits and thus much weaker separation in p-values. Instead, the observed strengthening separation is consistent with catastrophic-forgetting-driven reactivation of previously encoded information.

\subsection{Conditions for Success and Limitations}

Putting these observations together, CatShift is expected to work when the following conditions hold:

\begin{itemize}
\item \textbf{Sufficient encoding of the target dataset.} The model has meaningfully encoded or memorized the candidate dataset during pretraining.
\item \textbf{Geometric asymmetry.} Member data lies in sharper or more sensitive regions of the loss landscape compared to non-members.
\item \textbf{Effective parameter updates.} Fine-tuning induces non-trivial parameter changes rather than degenerate or null updates.
\end{itemize}

These conditions are commonly satisfied in overparameterized large language models trained on heterogeneous corpora, which explains the consistent empirical behavior observed across different architectures and model scales.

Conversely, CatShift may be less effective when these conditions are violated, for example when the target dataset is weakly represented in pretraining, when member and non-member data are highly indistinguishable in geometry, or when fine-tuning is severely constrained.

\section{Theoretical Analysis of CF-Induced Output Shifts}
\label{sec:theory}

In this section, we establish a theoretical framework to explain the mechanism underlying \textsc{CatShift}. We analyze why fine-tuning on member data triggers significantly larger output distribution shifts compared to non-member data. We model this phenomenon through the lens of loss landscape geometry, gradient dynamics, and information geometry.

Let $\theta^{(0)} \in \mathbb{R}^d$ denote the pre-trained parameters and $\theta^{(1)}$ denote the parameters after fine-tuning on a target subset $\mathcal{D}_s = \{(x_i, y_i)\}_{i=1}^N$. We quantify the output shift as the divergence between the predictive distributions $P(y|x; \theta^{(0)})$ and $P(y|x; \theta^{(1)})$.

\subsection{Loss Landscape Perspective: Curvature and Escape Dynamics}

We first examine the local geometry of the loss function $\mathcal{L}(\theta)$ around the pre-trained initialization $\theta^{(0)}$. Considering the second-order Taylor expansion for a sample $(x, y) \in \mathcal{D}_s$:

\begin{equation}
\begin{aligned}
\mathcal{L}(\theta) \approx\; & \mathcal{L}(\theta^{(0)}) 
+ (\theta - \theta^{(0)})^\top \nabla \mathcal{L}(\theta^{(0)}) \\
& + \frac{1}{2} (\theta - \theta^{(0)})^\top 
\mathbf{H}(\theta^{(0)}) (\theta - \theta^{(0)}),
\end{aligned}
\end{equation}
where $\mathbf{H}(\theta^{(0)}) = \nabla^2 \mathcal{L}(\theta^{(0)})$ is the Hessian matrix describing the local curvature.

\paragraph{Member Data (Sharp Basins).} For member data $x_{\text{mem}}$, the pre-trained model resides in a local minimum where the loss is low. However, it is well-established that deep models converge to \textit{sharp minima} for memorized data. This implies that the Hessian $\mathbf{H}(\theta^{(0)})$ possesses large positive eigenvalues in directions corresponding to memorized features. During fine-tuning, even a small gradient component aligned with these high-curvature directions causes the parameters to ``overshoot'' or escape the local basin rapidly. This \textit{over-correction} leads to a significant displacement in the parameter space relative to the basin's width.

\paragraph{Non-Member Data (Flat Plateaus).} Conversely, for non-member data $x_{\text{non}}$, the model typically resides on a high-loss plateau or a gentle slope where the curvature is low (small eigenvalues of $\mathbf{H}$). The optimization trajectory follows a smoother path, inducing diffuse updates that do not violently disrupt the pre-trained equilibrium.

\subsection{Gradient Alignment and Re-optimization Dynamics}

To understand the structural impact of the weight shift $\Delta \theta = \theta^{(1)} - \theta^{(0)}$, we analyze the alignment between the update vector and the pre-trained representations. 
Assuming a single step of gradient descent with learning rate $\eta$, the change in the model function $f(x; \theta)$ (pre-softmax logits) can be approximated in the Neural Tangent Kernel (NTK) regime:

\begin{equation}
\begin{aligned}
\Delta f(x) \approx\; & \nabla_{\theta} f(x; \theta^{(0)})^\top 
(\theta^{(1)} - \theta^{(0)}) \\
=\; & -\eta \left\| \nabla_{\theta} f(x; \theta^{(0)}) \right\|^2 
\frac{\partial \mathcal{L}}{\partial f}.
\end{aligned}
\end{equation}

\paragraph{Re-optimization vs. New Learning.}
For member data, the gradient $\nabla_{\theta} \mathcal{L}$ is highly aligned with the principal components of the pre-trained weights, effectively activating a ``Hessian Trace'' left during the pre-training phase. This triggers a \textit{structural re-optimization} of existing latent representations. In contrast, for non-members, the gradients are typically orthogonal to the dominant eigen-directions of the pre-trained manifold. This resembles a \textit{new learning} process (or feature appending) rather than a modification of core features, resulting in a less destructive interference with the model's output distribution.

\subsection{Information Geometry: Asymmetry in Output Shifts}

Finally, we map the parameter shift $\Delta \theta$ to the functional output shift using Information Geometry. We measure the output shift using the Kullback-Leibler (KL) divergence, approximated by the Fisher Information Matrix (FIM), $\mathcal{I}(\theta)$:

\begin{equation}
    D_{\text{KL}}(P_{\theta^{(0)}} || P_{\theta^{(1)}}) \approx \frac{1}{2} (\Delta \theta)^\top \mathcal{I}(\theta^{(0)}) (\Delta \theta),
\end{equation}
where $\mathcal{I}(\theta^{(0)}) = \mathbb{E}_{x \sim \mathcal{D}_s} \left[ \nabla \log P(y|x; \theta^{(0)}) \nabla \log P(y|x; \theta^{(0)})^\top \right]$.

This formulation reveals the fundamental asymmetry between members and non-members:

\begin{enumerate}
    \item \textbf{High Sensitivity for Members:} For member data, the model exhibits low entropy (high confidence), leading to an FIM $\mathcal{I}_{\text{mem}}$ with large spectral norms. The quadratic form $(\Delta \theta)^\top \mathcal{I}_{\text{mem}} (\Delta \theta)$ amplifies the parameter shift, resulting in a large divergence in probability space.
    \item \textbf{Low Sensitivity for Non-members:} For non-member data, the model has higher entropy, resulting in a flatter, more isotropic FIM $\mathcal{I}_{\text{non}}$. A similar magnitude of $\|\Delta \theta\|$ maps to a significantly smaller $D_{\text{KL}}$.
\end{enumerate}

Therefore, we conclude that the output shift is theoretically bounded by the curvature of the information manifold:
{\small
\begin{equation}
    (\Delta \theta_{\text{mem}})^\top \mathcal{I}_{\text{mem}} (\Delta \theta_{\text{mem}}) \gg (\Delta \theta_{\text{non}})^\top \mathcal{I}_{\text{non}} (\Delta \theta_{\text{non}}).
\end{equation}
}
This theoretical result underpins the empirical observation that membership inference can be reliably performed by measuring output distribution shifts under fine-tuning.

\section{Overall Algorithm}
\label{appendix:overall_algorithm}

\begin{algorithm}[ht]
\caption{Dataset Inference with \ourmethod{}}
\small
\label{alg:main}
\begin{algorithmic}[1]
\STATE \textbf{Input:} Target dataset $D_s = \{(x_i, y_i)\}_{i=1}^{N}$, target model $f^{(0)}$ with parameters $\theta^{(0)}$
\STATE \textbf{Output:} Likelihood of dataset $D_s$ being part of the training data
\STATE Split the target dataset $D_s$ into training and test subsets:
\[
D_s^{\text{train}}, D_s^{\text{test}} = \text{Split}(D_s)
\]
\STATE Construct completion function $c$.

\STATE Fine-tune the model $f^{(0)}$ on $D_s^{\text{train}}$ to obtain model $f^{(1)}$:
\[
\theta^{(1)} = \underset{\theta}{\mathrm{arg\,min}} \sum_{(x_i, y_i) \in D_s^{\text{train}}} \mathcal{L}(f^{(0)}(x_i), y_i)
\]
\FOR{each test sample $(x_i, y_i) \in D_s^{\text{test}}$}
    \STATE Compute top-1 completion from $f^{(0)}$: $\hat{y}_i^{(0)} = f^{(0)}(c(x_i))$
    \STATE Compute top-1 completion from $f^{(1)}$: $\hat{y}_i^{(1)} = f^{(1)}(c(x_i))$
    \STATE Compute similarity score $s_i = \text{Sim}(\hat{y}_i^{(0)}, \hat{y}_i^{(1)})$
\ENDFOR
\STATE Compute scores for the known non-member validation set $D_v$:
\[
s_j^v = \text{Sim}(\hat{y}_j^{(0)}, \hat{y}_j^{(1)}) \quad \text{for each } (x_j^v, y_j^v) \in D_v
\]
\STATE Apply Kolmogorov-Smirnov test: $p\text{-value} = \mathrm{KS}(S_s, S_v)$

\STATE \textbf{Decision:} 
\IF{$p\text{-value} < \alpha$}
    \STATE Reject the null hypothesis and conclude that $D_s$ is likely part of the model's training data
\ELSE
    \STATE Fail to reject the null hypothesis and conclude that $D_s$ is not part of the model's training data
\ENDIF
\end{algorithmic}
\end{algorithm}

Algorithm~\ref{alg:main} summarizes the procedure of \ourmethod{}. Given a candidate dataset, we first split it into training and test subsets and fine-tune the target model on the training portion using standard completion-style supervision. We then compare the model’s behavior before and after fine-tuning by computing top-1 completions for each test prompt under both the original and fine-tuned models. The similarity between these paired completions captures how fine-tuning alters the model’s outputs on the candidate data. To determine whether this shift reflects prior exposure or genuine adaptation, we contrast the resulting similarity scores with those obtained from a held-out validation dataset that is guaranteed not to appear in the model’s training corpus. Finally, we apply a Kolmogorov–Smirnov test to assess whether the two score distributions differ significantly. A low p-value indicates that fine-tuning induces a structured output shift distinct from that observed on non-member data, providing evidence that the candidate dataset was likely used during training.

\section{Comparison with PETAL}
\label{subsec:petal_comparison}

We further compare \ourmethod{} with a recent dataset inference method, PETAL~\cite{he2025towards}. The experiment is conducted on the Pythia-410M model across 22 subsets of the Pile dataset, following the same evaluation protocol as in \autoref{subsec:setup}. \autoref{tab:petal} reports the average performance.

\begin{table}[h]
\centering
\small
\begin{tabular}{lcc}
\toprule
\textbf{Method} & \textbf{AUC} & \textbf{TPR@5\%FPR} \\
\midrule
PDD~\cite{DBLP:conf/emnlp/Zhang0GRFC24} & 0.4865 & 0.0499 \\
LLM DI~\cite{maini2024llm} & 0.7890 & 0.5910 \\
PETAL~\cite{he2025towards} & 0.4650 & 0.0500 \\
\textbf{CatShift (Ours)} & \textbf{0.9790} & \textbf{0.9545} \\
\bottomrule
\end{tabular}
\caption{Comparison with PETAL on Pythia-410M across 22 Pile subsets.}
\label{tab:petal}
\end{table}

CatShift achieves substantially higher performance than PETAL and other baselines. In particular, CatShift improves AUC from 0.465 to 0.979 and achieves a TPR@5\%FPR of 0.9545, compared to 0.050 for PETAL. This gap is especially pronounced under the low-FPR regime, which is critical for practical auditing scenarios.

We note that the evaluation protocol does not place prior methods at a disadvantage. In particular, all methods are evaluated on the same model, datasets, and data splits, and prior approaches such as PETAL and LLM DI can fully utilize their intended signals under this setup. Despite this, CatShift consistently outperforms all baselines, suggesting that it provides a stronger and more reliable signal for dataset membership under token-only constraints.

\begin{figure*}[!t]
\centerline{\includegraphics[width=\linewidth]{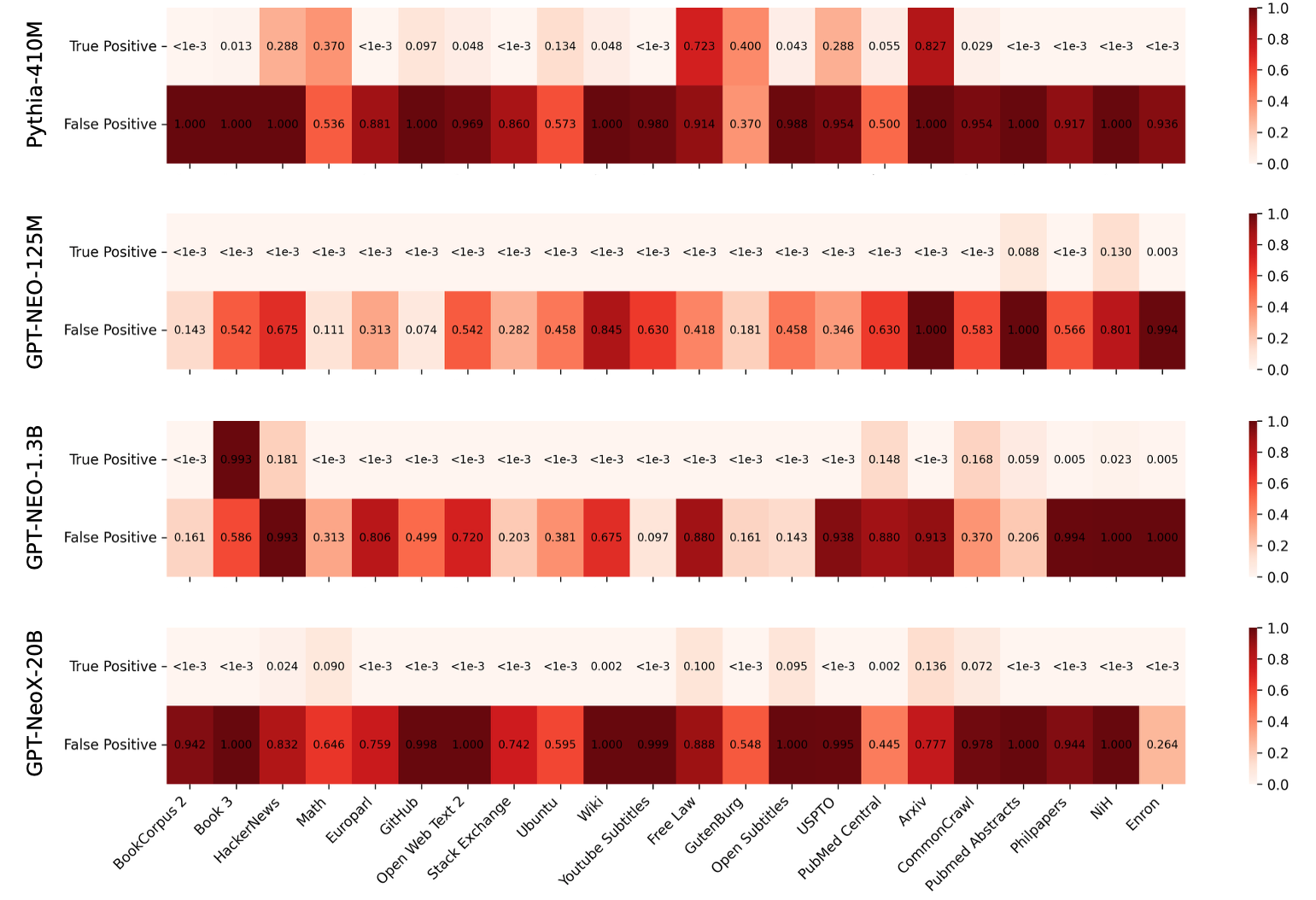}}
\caption{p-values of dataset inference by applying dataset inference to Pythia~\cite{DBLP:conf/icml/BidermanSABOHKP23} and GPT-Neo~\cite{black2021gptneo} models with 1000 data points. We can observe that \ourmethod{} can correctly distinguish train and validation splits of the PILE with low p-values.}
\label{fig:datasets}
\end{figure*}

\begin{figure*}[!t]
\centerline{\includegraphics[width=\linewidth]{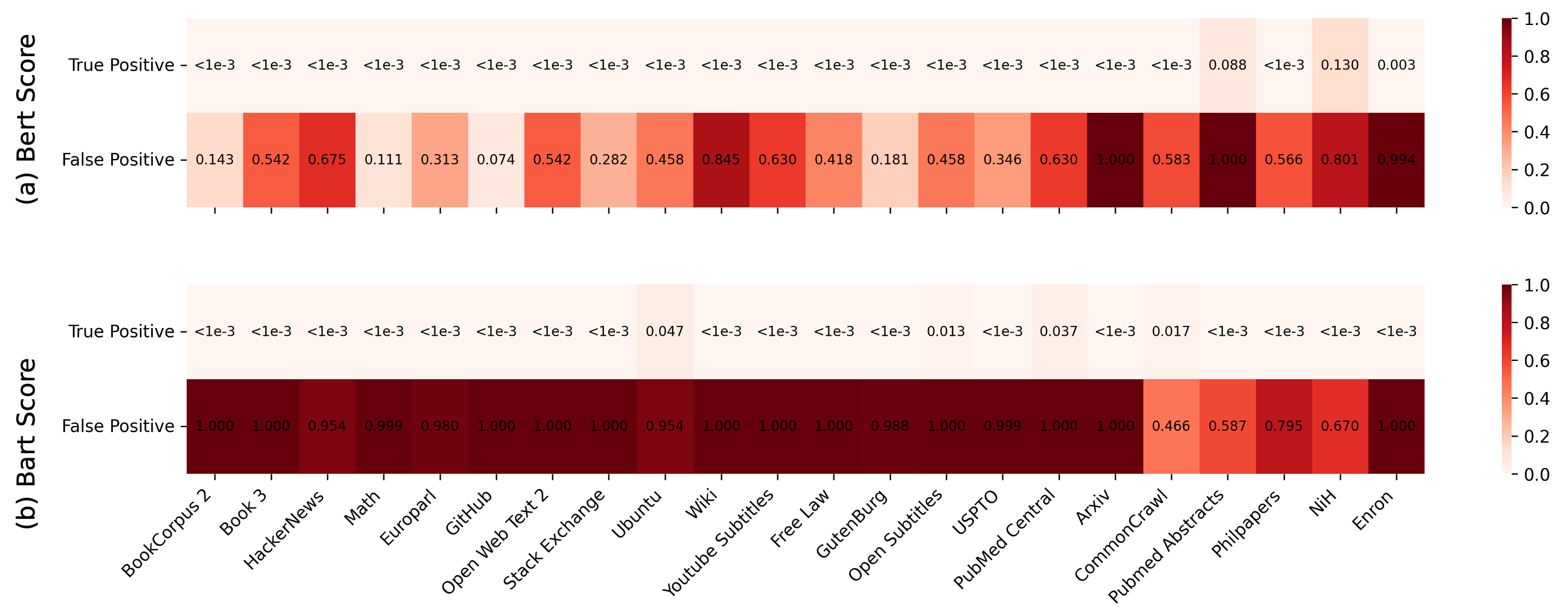}}
\caption{p-values of dataset inference on the GPT-Neo-125M model under different similarity scorers.}
\label{fig:bart}
\end{figure*}

\section{Impact of Similarity Scorer}

We study whether \ourmethod{} is sensitive to the choice of similarity scorer by comparing the default BERTScore with BartScore~\cite{DBLP:conf/nips/YuanNL21}, which is good at capturing semantic consistency. Experiments are conducted on the GPT-Neo-125M model. The results, shown in \autoref{fig:bart}, indicate that BartScore achieves comparable or slightly improved performance relative to BERTScore. This suggests that \ourmethod{} is not tightly coupled to a specific similarity metric and remains effective across different semantic scorers.

\section{In-Depth Analysis Across Subsets}
\label{subsec:Model}

Beyond aggregate performance, we conduct an in-depth analysis of \ourmethod{} at the level of individual PILE subsets to better understand how detection behavior varies across data sources and target models. While \autoref{tab:main} summarizes overall statistics across multiple architectures, it does not reveal which subsets drive false positives or false negatives. Here, we focus on subset-level p-value distributions to characterize these error patterns in detail.

\autoref{fig:datasets} shows the distribution of p-values for each PILE subset across four representative open-source LLMs. On Pythia-410M, \ourmethod{} exhibits extremely conservative behavior. At a threshold of 0.1, no non-member subset is misclassified, indicating strong control of false positives. However, this conservativeness leads to reduced sensitivity: seven member subsets produce p-values above the threshold, corresponding to a false negative rate of 31.8\%. This suggests that for smaller models, catastrophic-forgetting-induced shifts may be insufficiently strong or consistent across all training subsets, particularly those with weaker signals.

For the GPT-NEO family, the error profile shifts markedly. The method becomes substantially more sensitive, correctly identifying most member subsets. On GPT-NEO-125M, for example, only three member subsets yield p-values above $10^{-3}$, indicating that the majority of training sources are confidently detected. This gain in recall is accompanied by a moderate increase in false positives: at the 0.1 threshold, we observe one false positive on GPT-NEO-125M and GPT-NEO-1.3B, while none occur on GPT-NeoX-20B. These results indicate that more expressive models amplify output shifts on true members, but may also exaggerate spurious similarities for certain non-member subsets.

\para{Failure and Ambiguous-Case Analysis.}
A closer examination reveals that these errors are not uniformly distributed across subsets. Take Pythia-410M as an example, the highest false-negative rate consistently occurs on the \textit{arXiv} subset. A plausible explanation is that arXiv spans a wide range of scientific disciplines, resulting in substantial heterogeneity in topics, structures, and writing styles. This diversity weakens dataset-level linguistic regularities and reduces the likelihood that the model develops coherent memorization patterns tied to arXiv as a source. As a result, catastrophic forgetting induces weaker and less consistent output shifts, making true membership harder to detect.

In contrast, the highest false-positive rate is observed on the \textit{Gutenberg} subset. Gutenberg consists primarily of long-form literary works with relatively stable narrative structures and stylistic regularities. These properties can strengthen sequence-level memorization tendencies in LLMs and may also induce spurious familiarity effects or hallucinated recall during fine-tuning. Consequently, the detector may over-attribute these stylistic signals to training exposure, leading to inflated false-positive predictions.

\end{document}